\pdfoutput=1

\documentclass[a4paper,fleqn]{cas-dc}

\usepackage{amsmath}
\usepackage{amssymb}
\usepackage{array}

\usepackage[numbers]{natbib}
\usepackage{url}

\usepackage{subfig}
\usepackage{graphicx}

\usepackage{color}

\usepackage{listings} 
\definecolor{dkgreen}{rgb}{0,0.6,0}
\definecolor{gray}{rgb}{0.5,0.5,0.5}
\definecolor{mauve}{rgb}{0.58,0,0.82}

\usepackage{pifont}
\newcommand{\cmark}{\ding{51}}%
\newcommand{\xmark}{\ding{55}}%

\usepackage{lipsum}
\usepackage{hyperref}
\usepackage{enumitem}

\usepackage[frozencache,cachedir=.]{minted}
\setminted[r]{fontfamily=courier}
\usepackage{hhline}

\newcommand{\boldx}{\mathbf{x}}

\definecolor{orange}{rgb}{1,0.5,0}
\definecolor{redblue}{rgb}{1.0, 0.0, 1.0}
\definecolor{airforceblue}{rgb}{0.36, 0.54, 0.66}

\def\tsc#1{\csdef{#1}{\textsc{\lowercase{#1}}\xspace}}
\tsc{WGM}
\tsc{QE}
\tsc{EP}
\tsc{PMS}
\tsc{BEC}
\tsc{DE}

\begin{document}
\let\WriteBookmarks\relax
\def\floatpagepagefraction{1}
\def\textpagefraction{.001}
\shorttitle{pymoo: Multi-objective Optimization in Python}
\shortauthors{Julian Blank, Kalyanmoy Deb}

\title [mode = title]{pymoo: Multi-objective Optimization in Python}            



\address{Michigan State University, East Lansing, MI 48824, USA}

\author{Julian Blank}[type=editor,
                        orcid=0000-0002-2227-6476]
                        
\ead{blankjul@msu.edu}

\author{Kalyanmoy Deb}[type=editor,
                        orcid=0000-0001-7402-9939]
\ead{kdeb@msu.edu}


\begin{abstract}
Python has become the programming language of choice for research and industry projects related to data science, machine learning, and deep learning. Since optimization is an inherent part of these research fields, more optimization related frameworks have arisen in the past few years. Only a few of them support optimization of multiple conflicting objectives at a time, but do not provide comprehensive tools for a complete multi-objective optimization task. To address this issue, we have developed \texttt{pymoo}, a multi-objective optimization framework in Python. We provide a guide to getting started with our framework by demonstrating the implementation of an exemplary constrained multi-objective optimization scenario. Moreover, we give a high-level overview of the architecture of \texttt{pymoo} to show its capabilities followed by an explanation of each module and its corresponding sub-modules.
The implementations in our framework are customizable and algorithms can be modified/extended by supplying custom operators. Moreover, a variety of single, multi and many-objective test problems are provided and gradients can be retrieved by automatic differentiation out of the box. Also, \texttt{pymoo} addresses practical needs, such as the parallelization of function evaluations, methods to visualize low and high-dimensional spaces, and tools for multi-criteria decision making.
For more information about \texttt{pymoo}, readers are encouraged to visit: \url{https://pymoo.org}
\end{abstract}




\maketitle

\section{Introduction}

Optimization plays an essential role in many scientific areas, such as engineering, data analytics, and deep learning. These fields are fast-growing and their concepts are employed for various purposes, for instance gaining insights from a large data sets or fitting accurate prediction models. Whenever an algorithm has to handle a significantly large amount of data, an efficient implementation in a suitable programming language is important.
Python~\cite{python} has become the programming language of choice for the above mentioned research areas over the last few years, not only because it is easy to use but also good community support exists. Python is a high-level, cross-platform, and interpreted programming language that focuses on code readability. A large number of high-quality libraries are available and support for any kind of scientific computation is ensured. 
These characteristics make Python an appropriate tool for many research and industry projects where the investigations can be rather complex.
A fundamental principle of research is to ensure reproducibility of studies and to provide access to materials used in the research, whenever possible. In computer science, this translates to a sketch of an algorithm and the implementation itself.
However, the implementation of optimization algorithms can be challenging and specifically benchmarking is time-consuming. Having access to either a good collection of different source codes or a comprehensive library is time-saving and avoids an error-prone implementation from scratch.

To address this need for multi-objective optimization in Python, we introduce \texttt{pymoo}. The goal of our framework is not only to provide state of the art optimization algorithms, but also to cover different aspects related to the optimization process itself. 
We have implemented single, multi and many-objective test problems which can be used as a testbed for algorithms. In addition to the objective and constraint values of test problems, gradient information can be retrieved through automatic differentiation~\cite{automatic_differentiation}. Moreover, a parallelized evaluation of solutions can be implemented through vectorized computations, multi-threaded execution, and distributed computing.
Further, \texttt{pymoo} provides implementations of performance indicators to measure the quality of results obtained by a multi-objective optimization algorithm. Tools for an explorative analysis through visualization of lower and higher-dimensional data are available and multi-criteria decision making methods guide the selection of a single solution from a solution set based on preferences.

Our framework is designed to be extendable through of its modular implementation. For instance, a genetic algorithm is assembled in a plug-and-play manner by making use of specific sub-modules, such as initial sampling, mating selection, crossover, mutation and survival selection. Each sub-module takes care of an aspect independently and, therefore, variants of algorithms can be initiated by passing different combinations of sub-modules. This concept allows end-users to incorporate domain knowledge through custom implementations. For example, in an evolutionary algorithm a biased initial sampling module created with the knowledge of domain experts can guide the initial search.

Furthermore, we like to mention that our framework is well-documented with a large number of available code-snippets. 
We created a starter's guide for users to become familiar with our framework and to demonstrate its capabilities. As an example, it shows the optimization results of a bi-objective optimization problem with two constraints. An extract from the guide will be presented in this paper.
Moreover, we provide an explanation of each algorithm and source code to run it on a suitable optimization problem in our software documentation. Additionally, we show a definition of test problems and provide a plot of their fitness landscapes. The framework documentation is built using Sphinx~\cite{sphinx} and correctness of modules is ensured by automatic unit testing~\cite{python_unit_tests}. Most algorithms have been developed in collaboration with the second author and have been benchmarked  extensively against the original implementations. 

In the remainder of this paper, we first present related existing optimization frameworks in Python and in other programming languages. Then, we provide a guide to getting started with \texttt{pymoo} in Section~\ref{sec:getting_started} which covers the most important steps of our proposed framework. In Section~\ref{sec:architecture} we illustrate the framework architecture and the corresponding modules, such as problems, algorithms and related analytics. Each of the modules is then discussed separately in Sections~\ref{sec:problems} to \ref{sec:analytics}. Finally, concluding remarks are presented in Section~\ref{sec:conclusion}.

\section{Related Works}
In the last decades, various optimization frameworks in diverse programming languages were developed. However, some of them only partially cover multi-objective optimization. In general, the choice of a suitable framework for an optimization task is a multi-objective problem itself. Moreover, some criteria are rather subjective, for instance, the usability and extendibility of a framework and, therefore, the assessment regarding criteria as well as the decision making process differ from user to user. For example, one might have decided on a programming language first, either because of personal preference or a project constraint, and then search for a suitable framework. One might give more importance to the overall features of a framework, for example parallelization or visualization, over the programming language itself.
An overview of some existing multi-objective optimization frameworks in Python is listed in Table~\ref{tbl:frameworks}, each of which is described in the following. 

Recently, the well-known multi-objective optimization framework jMetal~\cite{jmetal} developed in Java~\cite{java} has been ported to a Python version, namely jMetalPy~\cite{jmetalpy}. The authors aim to further extend it and to make use of the full feature set of Python, for instance, data analysis and data visualization. In addition to traditional optimization algorithms, jMetalPy also offers methods for dynamic optimization. Moreover, the post analysis of performance metrics of an experiment with several independent runs is automated.

Parallel Global Multiobjective Optimizer, PyGMO~\cite{pygmo}, is an optimization library for the easy distribution of massive optimization tasks over multiple CPUs. It uses the generalized island-model paradigm for the coarse grained parallelization of optimization algorithms and, therefore, allows users to develop asynchronous and distributed algorithms.

Platypus~\cite{platypus} is a multi-objective optimization framework that offers implementations of state-of-the art algorithms. It enables users to create an experiment with various algorithms and provides post-analysis methods based on metrics and visualization.

A Distributed Evolutionary Algorithms in Python (DEAP) \cite{deap} is novel evolutionary computation framework for rapid prototyping and testing of ideas. Even though, DEAP does not focus on multi-objective optimization, however, due to the modularity and extendibility of the framework multi-objective algorithms can be developed. Moreover, parallelization and load-balancing tasks are supported out of the box.

Inspyred~\cite{inspyred} is a framework for creating bio-inspired computational intelligence algorithms in Python which is not focused on multi-objective algorithms directly, but on evolutionary computation in general. However, an example for NSGA-II~\cite{nsga2} is provided and other multi-objective algorithms can be implemented through the modular implementation of the framework.

If the search for frameworks is not limited to Python, other popular frameworks should be considered: PlatEMO~\cite{PlatEMO} in Matlab, MOEA~\cite{moea} and jMetal~\cite{jmetal} in Java, jMetalCpp~\cite{jMetalCpp} and PaGMO~\cite{pagmo} in \texttt{C++}. Of course this is not an exhaustive list and readers may search for other available options.

\begin{table}[]
\renewcommand{\arraystretch}{1.3}
\setlength\extrarowheight{2pt}
\caption{Multi-objective Optimization Frameworks in Python}
\centering
\begin{tabular}{|c|c|c|c|c|c|}
\hline
\multicolumn{1}{|p{0.9cm}}{\centering Name} & 
\multicolumn{1}{|p{0.7cm}}{\centering License} & 
\multicolumn{1}{|p{1.05cm}}{\centering Focus on\\ multi-objective} & 
\multicolumn{1}{|p{0.85cm}}{\centering Pure\\ Python} &
\multicolumn{1}{|p{0.85cm}}{\centering Vi\-su\-a\-li\-za\-tion} &
\multicolumn{1}{|p{0.92cm}|}{\centering Decision \\ Making}\\
\hline
jMetalPy & MIT & \cmark & \cmark & \cmark & \xmark \\
PyGMO & GPL-3.0 & \cmark & \xmark & \xmark & \xmark\\
Platypus & GPL-3.0 & \cmark & \cmark & \xmark & \xmark\\
DEAP &  LGPL-3.0 &  \xmark & \cmark & \xmark & \xmark \\
Inspyred & MIT & \xmark & \cmark & \xmark & \xmark \\
pymoo & Apache 2.0 & \cmark & \cmark & \cmark & \cmark \\ \hline
\end{tabular}
\label{tbl:frameworks}
\end{table}

\section{Getting Started \protect\footnote{All source codes in this paper are related to \texttt{pymoo} version 0.3.2. A getting started guide for upcoming versions can be found at pymoo.org.}}
\label{sec:getting_started}

In the following, we provide a starter's guide for \texttt{pymoo}. It covers the most important steps in an optimization scenario starting with the installation of the framework, defining an optimization problem, and the optimization procedure itself.

\subsection{Installation}

Our framework \texttt{pymoo} is available on PyPI~\cite{pypi} which is a central repository to make Python software package easily accessible. The framework can be installed by using the package manager:

\vspace{1mm}
\begin{minted}{bash}
   $ pip install -U pymoo
\end{minted}
\vspace{1mm}

Some components are available in Python and additionally in Cython~\cite{cython}. Cython allows developers to annotate existing Python code which is translated to \texttt{C} or \texttt{C++} programming languages. The translated files are compiled to a binary executable and can be used to speed up computations. 
During the installation of \texttt{pymoo}, attempts are made for compilation, however, if unsuccessful due to the lack of a suitable compiler or other reasons, the pure Python version is installed.
We would like to emphasize that the compilation is optional and all features are available without it.
More detail about the compilation and troubleshooting can be found in our installation guide online.

\subsection{Problem Definition}

In general, multi-objective optimization has several objective functions with subject to inequality and equality constraints to optimize\cite{multi_objective_book}. The goal is to find a set of solutions (variable vectors) that satisfy all constraints and are as good as possible regarding all its objectives values. The problem definition in its general form is given by:

\begin{align}
\label{eq:multi}
\begin{split}
\min          \quad& f_{m}(\boldx) \quad \quad \quad \quad m = 1,..,M,  \\[4pt]
\text{s.t.}   \quad& g_{j}(\boldx) \leq 0,  \quad \; \; \,  \quad j = 1,..,J, \\[2pt]
              \quad& h_{k}(\boldx) = 0,        \quad  \; \; \quad k = 1,..,K, \\[4pt]
              \quad& x_{i}^{L} \leq x_{i} \leq x_{i}^{U},  \quad i = 1,..,N. \\[2pt]
\end{split}
\end{align}

The formulation above defines a multi-objective optimization problem with $N$ variables, $M$ objectives, $J$ inequality, and $K$ equality constraints. Moreover, for each variable $x_i$, lower and upper variable boundaries ($x_i^L$ and $x_i^U$) are also defined.

In the following, we illustrate a bi-objective optimization problem with two constraints. 
\begin{align} 
\begin{split}
\min \;\; & f_1(x) = (x_1^2 + x_2^2), \\ 
\max \;\; & f_2(x) = -(x_1-1)^2 - x_2^2, \\[1mm] 
\text{s.t.} \;\; & g_1(x) = 2 \, (x_1 - 0.1) \, (x_1 - 0.9) \leq 0,\\ 
                       & g_2(x) = 20 \, (x_1 - 0.4) \, (x_1 - 0.6) \geq 0,\\[1mm] 
& -2 \leq x_1 \leq 2, \\
& -2 \leq x_2 \leq 2.
\end{split}
\label{eq:problem1}
\end{align}

\begin{figure}[]
\centering
\includegraphics[width=\linewidth]{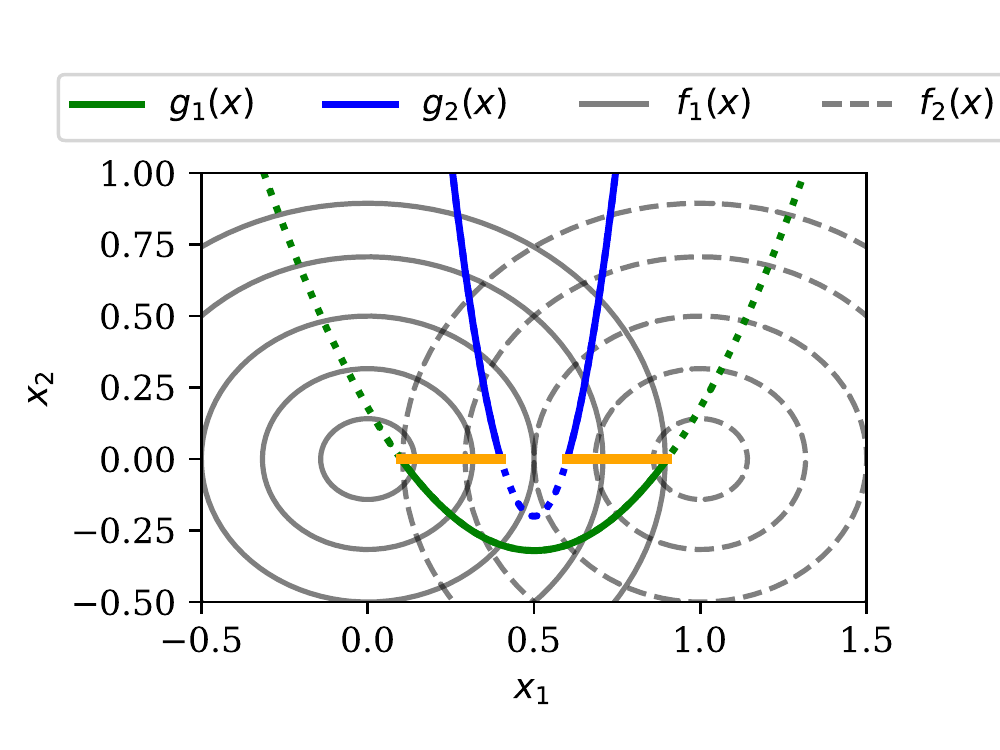}
\caption{Contour plot of the test problem~\ref{eq:problem1}.}
\label{fig:contour}
\end{figure}

It consists of two objectives ($M=2$) where $f_1(x)$ is minimized and $f_2(x)$ maximized. The optimization is with subject to two inequality constraints ($J=2$) where $g_1(x)$ is formulated as a less-than-equal-to and $g_2(x)$ as a greater-than-equal-to constraint. The problem is defined with respect to two variables ($N=2$), $x_1$ and $x_2$, which both are in the range $[-2,2]$. The problem does not contain any equality constraints ($K=0$). Contour plots of the objective functions are shown in Figure~\ref{fig:contour}. 
The contours of the objective function $f_1(x)$ are represented by solid lines and $f_2(x)$ by dashed lines. Constraints $g_1(x)$ and $g_2(x)$ are parabolas which intersect the $x_1$-axis at $(0.1, 0.9)$ and $(0.4, 0.6)$. The Pareto-optimal set is marked by a thick orange line. Through the combination of both constraints the Pareto-set is split into two parts.
Analytically, the Pareto-optimal set is  given by $PS = \{(x_1, x_2) \,|\, (0.1 \leq x_1 \leq 0.4) \lor (0.6 \leq x_1 \leq 0.9) \, \land \, x_2 = 0\}$ and the efficient-front by $f_2 = (\sqrt{f_1} - 1)^2$ where $f_1$ is defined in $[0.01,0.16]$ and $[0.36,0.81]$.  

In the following, we provide an example implementation of the problem formulation above using \texttt{pymoo}. We assume the reader is familiar with Python and has a fundamental knowledge of NumPy~\cite{numpy} which is utilized to deal with vector and matrix computations.

In \texttt{pymoo}, we consider pure minimization problems for optimization in all our modules. However, without loss of generality an objective which is supposed to be maximized, can be multiplied by $-1$ and be minimized \cite{deb-95}. Therefore, we minimize $-f_2(x)$ instead of maximizing $f_2(x)$ in our optimization problem. Furthermore, all constraint functions need to be formulated as a less-than-equal-to constraint. For this reason, $g_2(x)$ needs to be multiplied by $-1$ to flip the $\geq$ to a $\leq$ relation. We recommend the normalization of constraints to give equal importance to each of them. 
For $g_1(x)$, the constant `resource' value of the constraint is $2 \cdot (-0.1) \cdot (-0.9) = 0.18$ and for $g_2(x)$ it is $20 \cdot (-0.4) \cdot (-0.6) = 4.8$, respectively. We achieve normalization of constraints by dividing $g_1(x)$ and $g_2(x)$ by the corresponding constant \cite{debritu}.

Finally, the optimization problem to be optimized using \texttt{pymoo} is defined by:

\begin{align} 
\label{eq:getting_started_pymoo}
\begin{split}
\min \;\; & f_1(x) = (x_1^2 + x_2^2), \\ 
\min \;\; & f_2(x) = (x_1-1)^2 + x_2^2, \\[1mm] 
\text{s.t.} \;\; & g_1(x) = 2 \, (x_1 - 0.1) \, (x_1 - 0.9)  \, /  \,  0.18 \leq 0,\\ 
                       & g_2(x) = - 20 \, (x_1 - 0.4) \, (x_1 - 0.6) \,  /  \,  4.8 \leq 0,\\[1mm] 
& -2 \leq x_1 \leq 2, \\
& -2 \leq x_2 \leq 2.
\end{split}
\end{align}

Next, the derived problem formulation is implemented in Python. Each optimization problem in \texttt{pymoo} has to inherit from the \texttt{Problem} class. First, by calling the \texttt{super()} function the problem properties such as the number of variables (\texttt{n\textunderscore var}), objectives (\texttt{n\textunderscore obj}) and constraints (\texttt{n\textunderscore constr}) are initialized. Furthermore, lower (\texttt{xl}) and upper variables boundaries (\texttt{xu}) are supplied as a NumPy array. Additionally, the evaluation function $\texttt{\_evaluate}$ needs to be overwritten from the superclass. The method takes a two-dimensional NumPy array $\texttt{x}$ with $n$ rows and $m$ columns as an input. Each row represents an individual and each column an optimization variable. After doing the necessary calculations, the objective values are added to the dictionary $\texttt{out}$ with the key $\texttt{F}$ and the constraints with key $\texttt{G}$.

As mentioned above, \texttt{pymoo} utilizes NumPy~\cite{numpy} for most of its computations. To be able to retrieve gradients through automatic differentiation we are using a wrapper around NumPy called Autograd~\cite{autograd}. Note that this is not obligatory for a problem definition.

\begin{lstlisting}[language=Python]
import autograd.numpy as anp
from pymoo.model.problem import Problem

class MyProblem(Problem):

    def __init__(self):
        super().__init__(n_var=2, 
                         n_obj=2, 
                         n_constr=2, 
                         xl=anp.array([-2,-2]), 
                         xu=anp.array([2,2]))

    def _evaluate(self, x, out, *args, **kwargs):
        f1 = x[:,0]**2 + x[:,1]**2
        f2 = (x[:,0]-1)**2 + x[:,1]**2
        
        g1 = 2*(x[:, 0]-0.1) * (x[:, 0]-0.9) / 0.18
        g2 = - 20*(x[:, 0]-0.4) * (x[:, 0]-0.6) / 4.8
        
        out["F"] = anp.column_stack([f1, f2])
        out["G"] = anp.column_stack([g1, g2])

\end{lstlisting}

\subsection{Algorithm Initialization}

Next, we need to initialize a method to optimize the problem.
In \texttt{pymoo}, an algorithm object needs to be created for optimization. For each of the algorithms an API documentation is available and through supplying different parameters, algorithms can be customized in a plug-and-play manner.
In general, the choice of a suitable algorithm for optimization problems is a challenge itself. Whenever problem characteristics are known beforehand we recommended using those through customized operators.
However, in our case the optimization problem is rather simple, but the aspect of having two objectives and two constraints should be considered. For this reason, we decided to use NSGA-II~\cite{nsga2} with its default configuration with minor modifications. We chose a population size of $40$, but instead of generating the same number of offsprings, we create only $10$ each generation. This is a steady-state variant of NSGA-II and it is likely to improve the convergence property for rather simple optimization problems without much difficulties, such as the existence of local Pareto-fronts.
Moreover, we enable a duplicate check which makes sure that the mating produces offsprings which are different with respect to themselves and also from the existing population regarding their variable vectors. To illustrate the customization aspect, we listed the other unmodified default operators in the code-snippet below. The constructor of \texttt{NSGA2} is called with the supplied parameters and returns an initialized algorithm object.

\begin{lstlisting}[language=Python]
from pymoo.algorithms.nsga2 import NSGA2
from pymoo.factory import get_sampling, get_crossover, get_mutation

algorithm = NSGA2(
    pop_size=40,
    n_offsprings=10,
    sampling=get_sampling("real_random"),
    crossover=get_crossover("real_sbx", prob=0.9, eta=15),
    mutation=get_mutation("real_pm", eta=20),
    eliminate_duplicates=True
)

\end{lstlisting}

\subsection{Optimization}

Next, we use the initialized algorithm object to optimize the defined problem. Therefore, the \texttt{minimize} function with both instances \texttt{problem} and \texttt{algorithm} as parameters is called. Moreover, we supply the termination criterion of running the algorithm for $40$ generations which will result in $40 + 40 \times 10 = 440$ function evaluations.
In addition, we define a random seed to ensure reproducibility and enable the verbose flag to see printouts for each generation. 
The method returns a \texttt{Result} object which contains the non-dominated set of solutions found by the algorithm.

\begin{lstlisting}[language=Python]
from pymoo.optimize import minimize

res = minimize(MyProblem(),
               algorithm,
               ('n_gen', 40),
               seed=1,
               verbose=True)

\end{lstlisting}

\begin{figure} 
    \centering
  \subfloat[Design Space\label{fig:exmple_result_x}]{%
       \includegraphics[width=\linewidth]{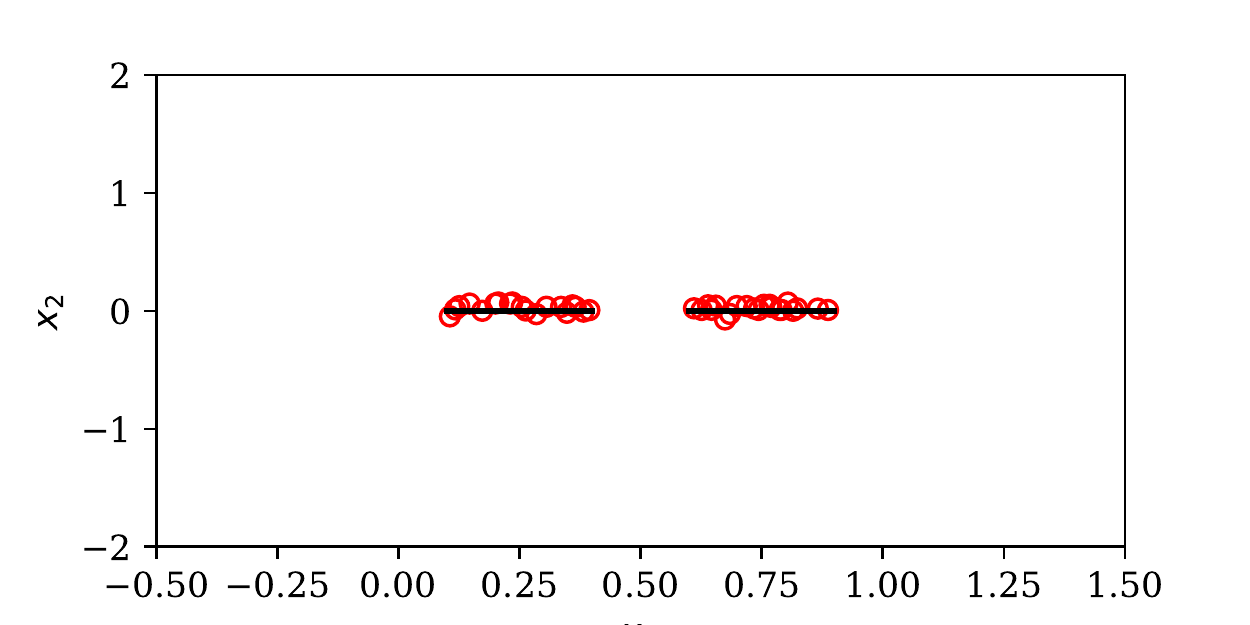}}
       \\
  \subfloat[Objective Space\label{fig:exmple_result_f}]{%
        \includegraphics[width=\linewidth]{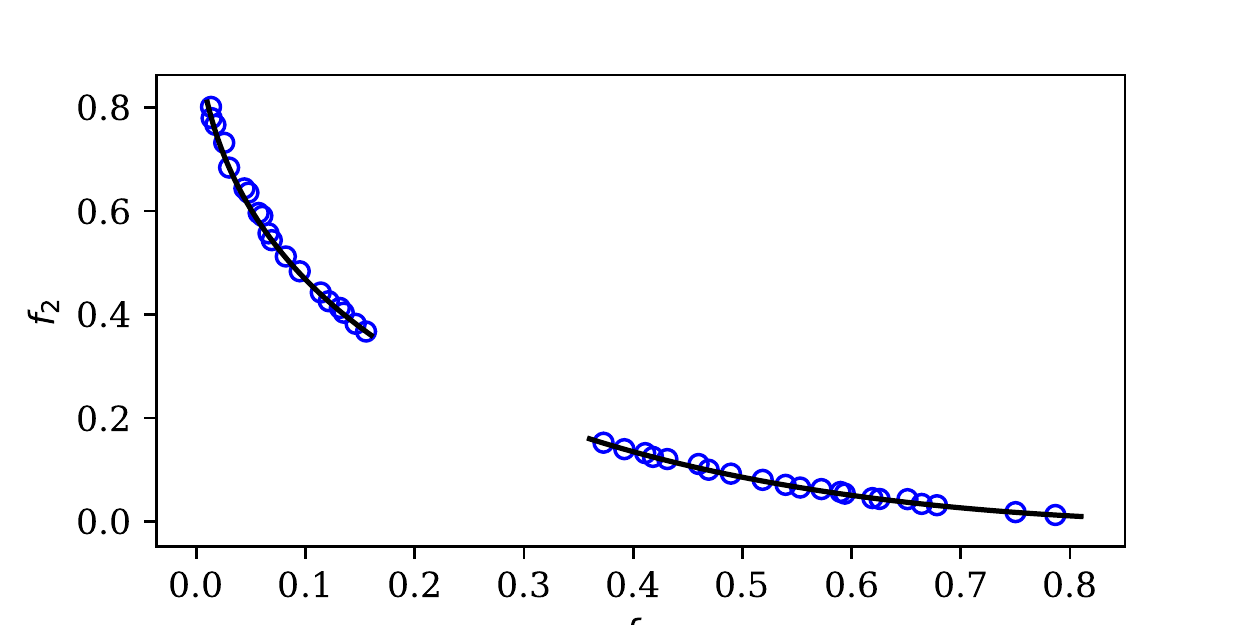}}
  \caption{Result of the Getting Started Optimization}
    \label{fig:example_result}
\end{figure}

The optimization results are illustrated in Figure~\ref{fig:example_result} where the design space is shown in Figure~\ref{fig:exmple_result_x} and in the objective space in Figure~\ref{fig:exmple_result_f}. The solid line represents the analytically derived Pareto set and front in the corresponding space and the circles solutions found by the algorithm. It can be observed that the algorithm was able to converge and a set of nearly-optimal solutions was obtained. Some additional post-processing steps and more details about other aspects of the optimization procedure can be found in the remainder of this paper and in our software documentation.

The starters guide showed the steps starting from the installation up to solving an optimization problem. The investigation of a constrained bi-objective problem demonstrated the basic procedure in an optimization scenario.

\section{Architecture}
\label{sec:architecture}

\begin{figure*}[]
\centering
\includegraphics[width=\linewidth]{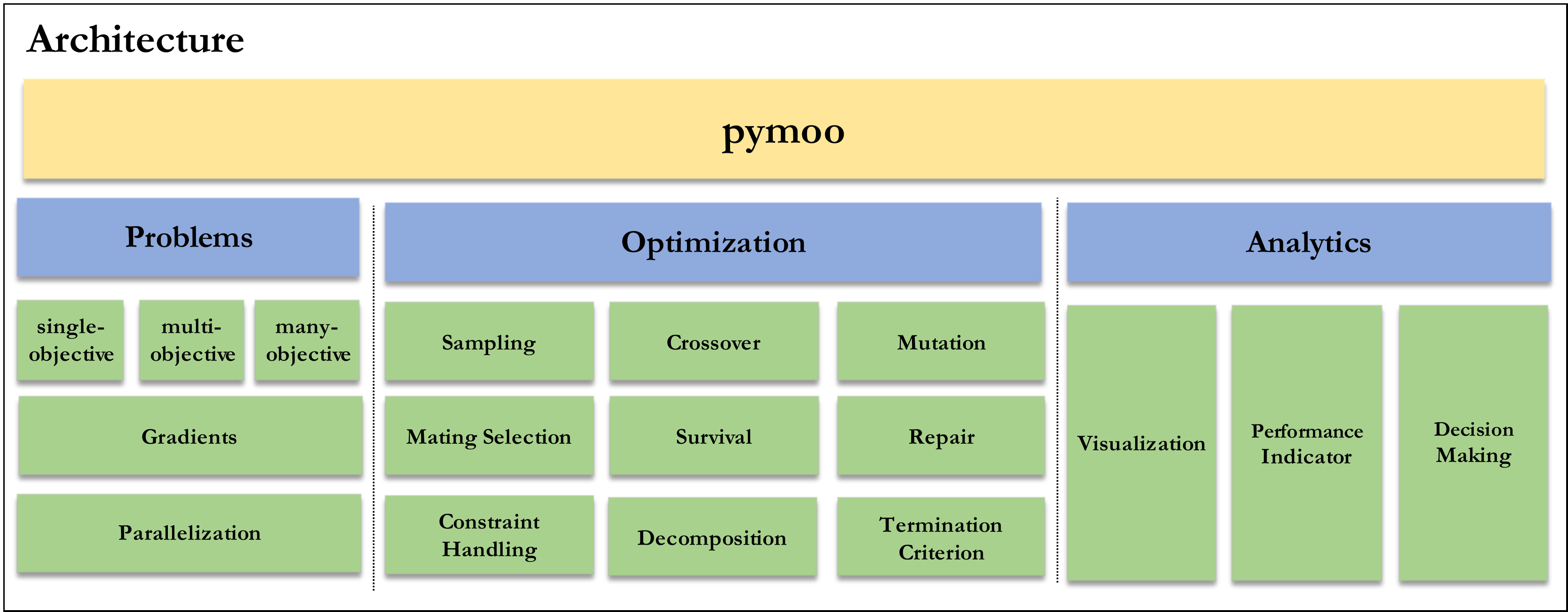}
\caption{Architecture of \texttt{pymoo}.}
\label{fig:architecture}
\end{figure*}

Software architecture is fundamentally important to keep source code organized. On the one hand, it helps developers and users to get an overview of existing classes, and on the other hand, it allows flexibility and extendibility by adding new modules. Figure~\ref{fig:architecture} visualizes the architecture of \texttt{pymoo}. The first level of abstraction consists of the optimization problems, algorithms and analytics. Each of the modules can be categorized into more detail and consists of multiple sub-modules.

\begin{enumerate}[label=(\roman*)]
\item \textbf{Problems:} Optimization problems in our framework are categorized into single, multi, and many-objective test problems. Gradients are available through automatic differentiation and parallelization can be implemented by using a variety of techniques.

\item \textbf{Optimization:} Since most of the algorithms are based on evolutionary computations, operators such as sampling, mating selection, crossover and mutation have to be chosen or implemented. Furthermore, because many problems in practice have one or more constraints, a methodology for handling those must be incorporated. Some algorithms are based on decomposition which splits the multi-objective problem into many single-objective problems. Moreover, when the algorithm is used to solve the problem, a termination criterion must be defined either explicitly or implicitly by the implementation of the algorithm. 

\item \textbf{Analytics:} During and after an optimization run analytics support the understanding of data. First, intuitively the design space, objective space, or other metrics can be explored through visualization. Moreover, to measure the convergence and/or diversity of a Pareto-optimal set performance indicators can be used. To support the decision making process either through finding points close to the area of interest in the objective space or high trade-off solutions. This can be applied either during an optimization run to mimic interactive optimization or as a post analysis.

\end{enumerate}

In the remainder of the paper, we will discuss each of the modules mentioned in more detail. 

\section{Problems}
\label{sec:problems}

It is common practice for researchers to evaluate the performance of algorithms on a variety of test problems. Since we know no single-best algorithm for all arbitrary optimization problems exist~\cite{nfl}, this helps to identify problem classes where the algorithm is suitable.
Therefore, a collection of test problems with different numbers of variables, objectives or constraints and alternating complexity becomes handy for algorithm development.
Moreover, in a multi-objective context, test problems with different Pareto-front shapes or varying variable density close to the optimal region are of interest.

\subsection{Implementations}

In our framework, we categorize test problems regarding the number of objectives: single-objective (1 objective), multi-objective (2 or 3 objectives) and many-objective (more than 3 objectives). Test problems implemented in \texttt{pymoo} are listed in Table~\ref{tbl:problems}. For each problem the number of variables, objectives, and constraints are indicated. If the test problem is scalable to any of the paramaters, we label the problem with \textit{(s)}. If the problem is scalable, but a default number was original proposed we indicate that with surrounding brackets. In case the category does not apply, for example because we refer to a test problem family with several functions, we use \textit{($\cdot$)}.

\begin{table}[]
\footnotesize
\renewcommand{\arraystretch}{1.3}
\caption{Multi-objective Optimization Test problems.}
\centering
\begin{tabular}{cccc}
\textbf{Problem} & \textbf{Variables} & \textbf{Objectives} & \textbf{Constraints}\\[1mm] \hline
\multicolumn{4}{c}{\textbf{Single-Objective}} \\ \hline\hline
Ackley&(s)&1&-\\
Cantilevered Beams&4&1&2\\
Griewank&(s)&1&-\\
Himmelblau&2&1&-\\
Knapsack&(s)&1&1\\
Pressure Vessel&4&1&4\\
Rastrigin&(s)&1&-\\
Rosenbrock&(s)&1&-\\
Schwefel&(s)&1&-\\
Sphere&(s)&1&-\\
Zakharov&(s)&1&-\\ 
G1-9&($\cdot$)&($\cdot$)&($\cdot$)\\ \hline\hline
\multicolumn{4}{c}{\textbf{Multi-Objective}} \\ \hline\hline
BNH&2&2&2\\
Carside&7&3&10\\
Kursawe&3&2&-\\
OSY&6&2&6\\
TNK&2&2&2\\
Truss2D&3&2&1\\
Welded Beam&4&2&4\\
CTP1-8&(s)&2&(s)\\
ZDT1-3&(30)&2&-\\
ZDT4&(10)&2&-\\
ZDT5&(80)&2&-\\
ZDT6&(10)&2&-\\ \hline\hline
\multicolumn{4}{c}{\textbf{Many-Objective}} \\ \hline\hline
DTLZ 1-7&(s)&(s)&-\\
CDTLZ&(s)&(s)&-\\ 
$\text{DTLZ1}^{-1}$&(s)&(s)&-\\ 
SDTLZ&(s)&(s)&-\\  \hline
\end{tabular}
\label{tbl:problems}
\end{table}

The implementations in \texttt{pymoo} let end-users define what values of the corresponding problem should be returned. 
On an implementation level, the \texttt{evaluate} function of a \texttt{Problem} instance takes a list \texttt{return\_value\_of} which contains the type of values being returned. By default the objective values \texttt{"F"} and if the problem has constraints the constraint violation \texttt{"CV"} are included. The constraint function values can be returned independently by adding \texttt{"G"}. This gives developers the flexibility to receive the values that are needed for their methods.

\subsection{Gradients}
\label{sec:gradients}
All our test problems are implemented using Autograd~\cite{autograd}. Therefore, automatic differentiation is supported out of the box. We have shown in Section~\ref{sec:getting_started} how a new optimization problem is defined. 

If gradients are desired to be calculated the prefix \texttt{"d"} needs to be added to the corresponding value of the \texttt{return\_value\_of} list. For instance to ask for the objective values and its gradients \texttt{return\_value\_of = ["F", "dF"]}.

Let us consider the problem we have implemented shown in Equation~\ref{eq:getting_started_pymoo}. The derivation of the objective functions $F$ with respect to each variable is given by:

\begin{gather}
\nabla F
 =
  \begin{bmatrix}
   2 x_1 &  2 x_2\\
   2 (x_1-1) & 2 x_2
   \end{bmatrix}.
\end{gather}

The gradients at the point $[0.1, 0.2]$ are calculated by: 

\begin{lstlisting}[language=Python]
F, dF = problem.evaluate(np.array([0.1, 0.2]), 
                         return_values_of=["F", "dF"])
\end{lstlisting}

returns the following output 

\begin{lstlisting}[language=Bash]
F  <- [0.05, 0.85]
dF <- [[ 0.2,  0.4],
       [-1.8,  0.4]]
\end{lstlisting}

It can easily be verified that the values are matching with the analytic gradient derivation. 
The gradients for the constraint functions can be calculated accordingly by adding \texttt{"dG"} to the \texttt{return\_value\_of} list.

\subsection{Parallelization}
\label{sec:parallelization}

If evaluation functions are computationally expensive, a serialized evaluation of a set of solutions can become the bottleneck of the overall optimization procedure.
For this reason, parallelization is desired for an use of existing computational resources more efficiently and distribute long-running calculations.
In \texttt{pymoo}, the evaluation function receives a set of solutions if the algorithm is utilizing a population. This empowers the user to implement any kind of parallelization as long as the objective values for all solutions are written as an output when the evaluation function terminates.
In our framework, a couple of possibilities to implement parallelization exist:

\begin{enumerate}[label=(\roman*)]
\item \textbf{Vectorized Evaluation:} 
A common technique to parallelize evaluations is to use matrices where each row represents a solution. Therefore, a vectorized evaluation refers to a column which includes the variables of all solutions. By using vectors the objective values of all solutions are calculated at once.
The code-snippet of the example problem in Section~\ref{sec:getting_started} shows such an implementation using NumPy~\cite{numpy}. 
To run calculations on a GPU, implementing support for PyTorch~\cite{pytorch} tensors can be done with little overhead given suitable hardware and correctly installed drivers.

\item \textbf{Threaded Loop-wise Evaluation:} If the function evaluation should occur independently, a \texttt{for} loop can be used to set the values. By default the evaluation is serialized and no calculations occur in parallel. By providing a keyword to the evaluation function, \texttt{pymoo} spawns a thread for each evaluation and manages those by using the default thread pool implementation in Python. This behaviour can be implemented out of the box and the number of parallel threads can be modified.

\item \textbf{Distributed Evaluation:}
If the evaluation should not be limited to a single machine, the evaluation itself can be distributed to several workers or a whole cluster. We recommend using Dask~\cite{dask} which enables distributed computations on different levels. For instance, the matrix operation itself can be distributed or a whole function can be outsourced.
Similar to the loop wise evaluation each individual can be evaluate element-wise by sending it to a worker.
\end{enumerate}

\section{Optimization Module}

The optimization module provides different kinds of sub-modules to be used in algorithms. Some of them are more of a generic nature, such as decomposition and termination criterion, and others are more related to evolutionary computing. By assembling those modules together algorithms are built. 

\subsection{Algorithms}
Available algorithm implementations in \texttt{pymoo} are listed in Table~\ref{tbl:algorithms}. Compared to other optimization frameworks the list of algorithms may look rather short, however, each algorithm is customizable and variants can be initialized with different parameters. For instance, a Steady-State NSGA-II~\cite{nsga2_steady_state} can be initialized by setting the number of offspring to $1$. This can be achieved by supplying this as a parameter in the initialization method as shown in Section~\ref{sec:getting_started}.

\begin{table}[]
\footnotesize
\renewcommand{\arraystretch}{1.3}
\caption{Multi-objective Optimization Algorithms.}
\centering
\begin{tabular}{llll}
Algorithm & Reference\\\hline
GA & \\
DE & \cite{de}\\ \hline
NSGA-II & \cite{nsga2}\\
RNSGA-II & \cite{rnsga2}\\
NSGA-III & \cite{nsga3-part1, nsga3-part2, nsga3-norm}\\
UNSGA-III & \cite{unsga3}\\
RNSGA-III & \cite{rnsga3}\\
MOEAD & \cite{moead} \\ \hline
\end{tabular}
\label{tbl:algorithms}
\end{table}

\subsection{Operators}

\begin{figure} 
    \centering
  \subfloat[\quad One Point\label{fig:one_point}]{%
       \includegraphics[width=0.45\linewidth]{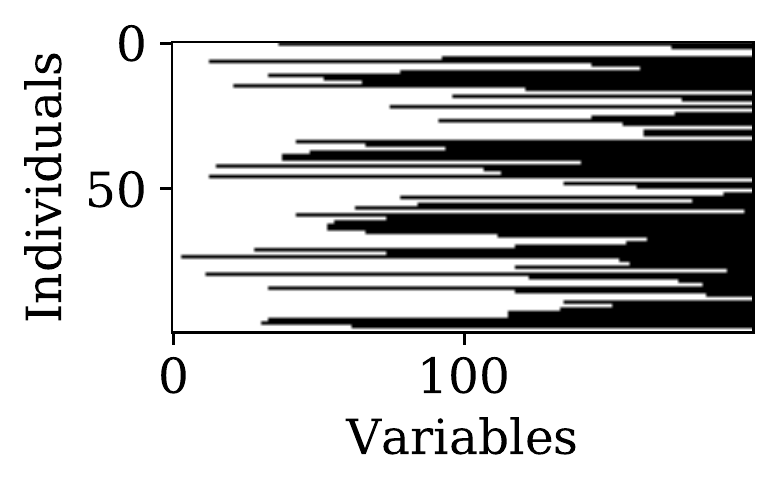}}
    \hfill
  \subfloat[Two Point\label{fig_two_point}]{%
        \includegraphics[width=0.45\linewidth]{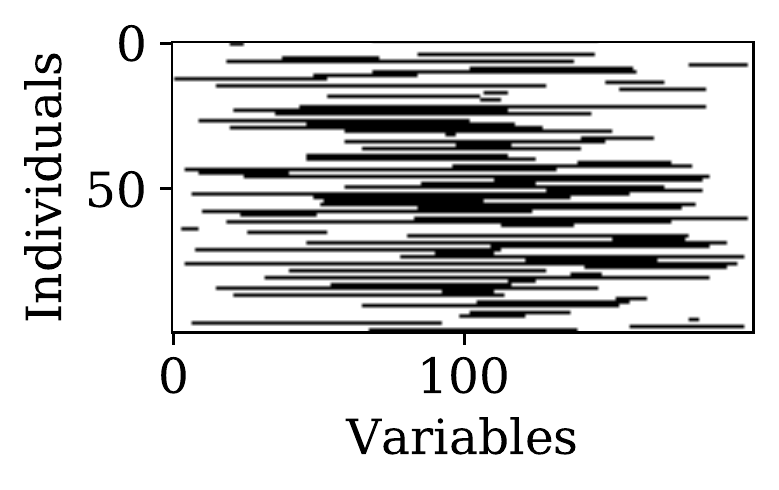}}
    \\
  \subfloat[UX\label{fig:ux}]{%
        \includegraphics[width=0.45\linewidth]{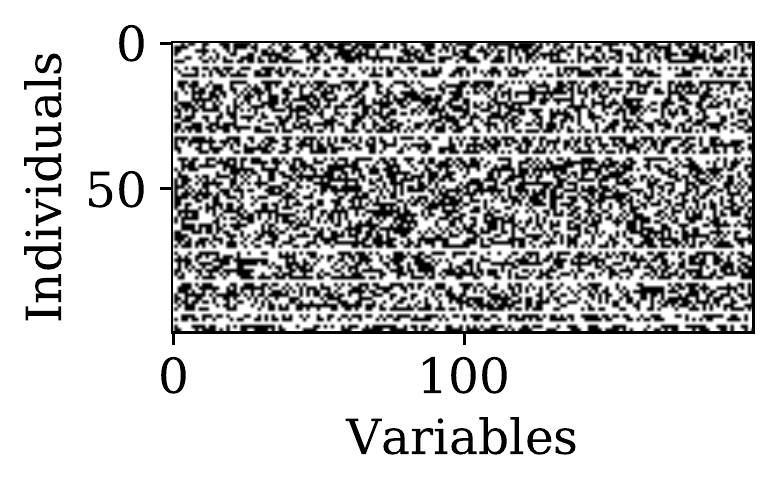}}
    \hfill
  \subfloat[HUX\label{fig:hux}]{%
        \includegraphics[width=0.45\linewidth]{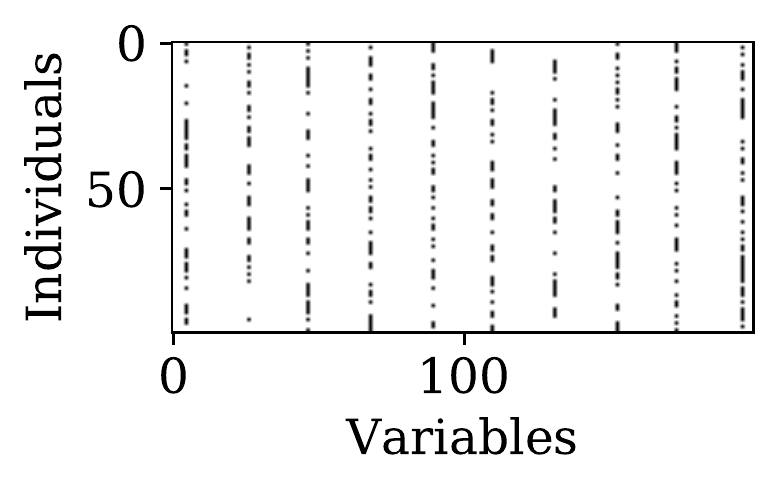}}
    \\
  \subfloat[SBX (real, eta=0.8)\label{fig:sbx_real}]{%
        \includegraphics[width=0.45\linewidth]{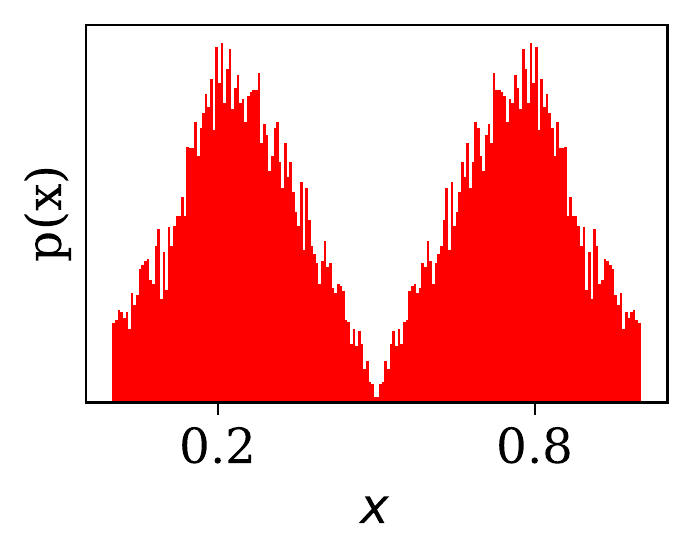}}
    \hfill
  \subfloat[SBX (int, eta=3)\label{fig:sbx_int}]{%
        \includegraphics[width=0.45\linewidth]{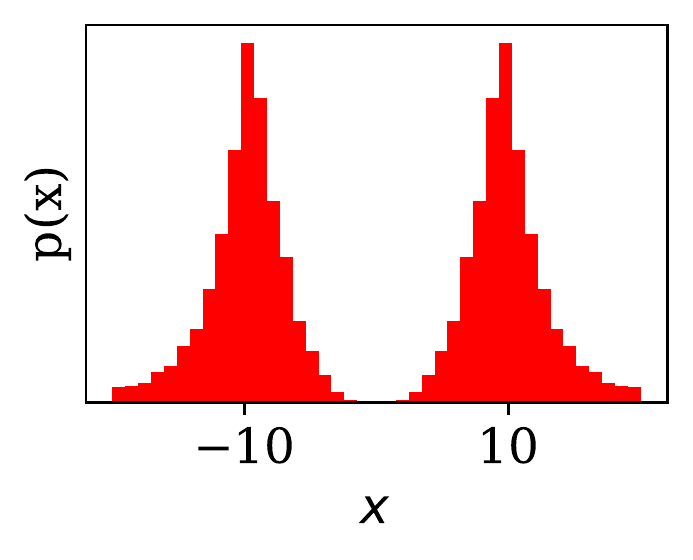}}
  \caption{Illustration of some crossover operators for different variables types.}
  \label{fig:crossover} 
\end{figure}

The following evolutionary operators are available:

\begin{enumerate}[label=(\roman*)]
\item \textbf{Sampling:} The initial population is mostly based on sampling. In some cases it is created through domain knowledge and/or some solutions are already evaluated, they can directly be used as an initial population. Otherwise, it can be sampled randomly for real, integer, or binary variables. Additionally, Latin-Hypercube Sampling~\cite{lhs} can be used for real variables.

\item \textbf{Crossover:} A variety of crossover operators for different type of variables are implemented. In Figure~\ref{fig:crossover} some of them are presented. 
Figures~\ref{fig:one_point}-~\ref{fig:hux} help to visualize the information exchange in a crossover with two parents being involved. Each row represents an offspring and each column a variable. The corresponding boxes indicate whether the values of the offspring are inherited from the first or from the second parent. For one and two point crossovers it can be observed that either one or two cuts in the variable sequence exist. Contrarily, the Uniform Crossover (UX) does not have any clear pattern, because each variable is chosen randomly either from the first or from the second parent. For the Half Uniform Crossover (HUX) half of the variables, which are different, are exchanged. For the purpose of illustration, we have created two parents that have different values in $10$ different positions. 
For real variables, Simulated Binary Crossover~\cite{sbx} is known to be an efficient crossover. It mimics the crossover of binary encoded variables. In Figure~\ref{fig:sbx_real} the probability distribution when the parents $x_1=0.2$ and $x_2=0.8$ where $x_i \in [0, 1]$ with $\eta = 0.8$ are recombined is shown. Analogously, in case of integer variables we subtract $0.5$ from the lower and add $0.5-\epsilon$ to the upper bound before applying the crossover and round to the nearest integer afterwards (see Figure~\ref{fig:sbx_int}).

\item \textbf{Mutation:} For real and integer variables Polynomial Mutation~\cite{sbx} and for binary variables Bitflip mutation is provided.
\end{enumerate}

Different problems require different type of operators. In practice, if a problem is supposed to be solved repeatedly, it makes sense to customize the evolutionary operators to improve the convergence of the algorithm. Moreover, for custom variable types, for instance trees or mixed variables, custom operators can be implemented easily and called by algorithm class. Our software documentation contains examples for custom modules, operators and variable types.

\subsection{Termination Criterion}
For every algorithm it must be determined when it should terminate a run. This can be simply based on a predefined number of function evaluations, iterations, or more advanced criteria such as the change of a performance metric over time.
For example, we have implemented a termination criterion based on the design space difference between generations. To make the termination criterion more robust the last $k$ generations are considered. The largest movement from a solution to its closest neighbour is tracked across generation and whenever it is below a certain threshold the algorithm is considered to have converged. Analogously, the movement in the objective space can be chosen for termination in \texttt{pymoo}.

\subsection{Decomposition}
\label{sec:decomp}
Decomposition transforms multi-objective problems into many single-objective optimization problems~\cite{decomp}. Such a technique can be either embedded in a multi-objective algorithm and solved simultaneously or independently using a single-objective optimizer. Some decomposition methods are based on the \textit{lp}-metrics with different $p$ values.
For instance, a naive but frequently used decomposition approach is the Weighted-Sum Method ($p=1$), which is known to be not able to converge to the non-convex part of a Pareto-front~\cite{multi_objective_book}. Moreover, instead of summing values, Tchebysheff Method ($p=\infty$) considers only the maximum value of the difference between the ideal point and a solution.
Similarly, the Achievement Scalarization Function (ASF)~\cite{asf} and a modified version Augmented Achievement Scalarization Function (AASF)~\cite{aasf} use the maximum of all differences.
Furthermore, Penalty Boundary Intersection (PBI)~\cite{moead} is calculated by a weighted sum of the norm of the projection of a point onto the reference direction and the perpendicular distance.
Also it is worth to note that normalization is essential for any kind of decomposition.
All decomposition techniques mentioned above are implemented in \texttt{pymoo}.

\section{Analytics}
\label{sec:analytics}

\subsection{Performance Indicators}
\label{sec:pi}

For single-objective optimization algorithms the comparison regarding performance is rather simple because each optimization run results in a single best solution. In multi-objective optimization, however, each run returns a non-dominated set of solutions. To compare sets of solutions various performance indicators have been proposed in the past~\cite{pi}. In \texttt{pymoo} most commonly used performance indicators are described:

\begin{enumerate}[label=(\roman*)]
\item \textbf{GD/IGD:} Given the Pareto-front \texttt{PF} the deviation between the non-dominated set \texttt{S} found by the algorithm and the optimum can be measured. Following this principle, Generational Distance (GD) Indicator~\cite{gd} calculates the average euclidean distance in the objective space from each solution in \texttt{S} to the closest solution in \texttt{PF}. 
This measures the convergence of \texttt{S}, but does not indicate whether a good diversity on the Pareto-front has been reached.
Similarly, Inverted Generational Distance (IGD) Indicator~\cite{gd} measures the average Euclidean distance in the objective space from each solution in \texttt{PF} to the closest solution in \texttt{S}. The Pareto-front as a whole needs to be covered by solutions from \texttt{S} to minimize the performance metric. However, IGD is known to be not Pareto compliant~\cite{igd_plus}.
\item \textbf{GD+/IGD+:} A variation of GD and IGD has been proposed in~\cite{igd_plus}. The Euclidean distance is replaced by a distance measure that takes the dominance relation into account. The authors show that IGD+ is weakly Pareto compliant.
\item \textbf{Hypervolume:} Moreover, the dominated portion of the objective space can be used to measure the quality of non-dominated solutions~\cite{hv}. Instead of the Pareto-front a reference point needs to be provided. It has been shown that Hypervolume is Pareto compliant~\cite{hv_Pareto_compliant}. Because the performance metric becomes computationally expensive in higher dimensional spaces  the exact measure becomes intractable. However, we plan to include some proposed approximation methods in the near future. 
\end{enumerate}

Performance indicators are used to compare existing algorithms. Moreover, the development of new algorithms can be driven by the goodness of different metrics itself.

\subsection{Visualization}
The visualization of intermediate steps or the final result is inevitable. In multi and many-objective optimization visualization of the objective space is of interest especially, and focuses on visualizing trade-offs between solutions.
Depending on the dimensionality different types of plots are suitable to represent a single or a set of solutions. In \texttt{pymoo} the implemented visualizations wrap around the well-known plotting library in Python Matplotlib~\cite{matplotlib}. Keyword arguments provided by Matplotlib itself are still available which allows to modify for instance the color, thickness, opacity of lines, points or other shapes. Therefore, all visualization techniques are customizable and extendable.

\begin{figure*} 
    \centering
  \subfloat[\quad Scatter 2D\label{fig:viz_scatter}]{%
       \includegraphics[width=0.275\linewidth]{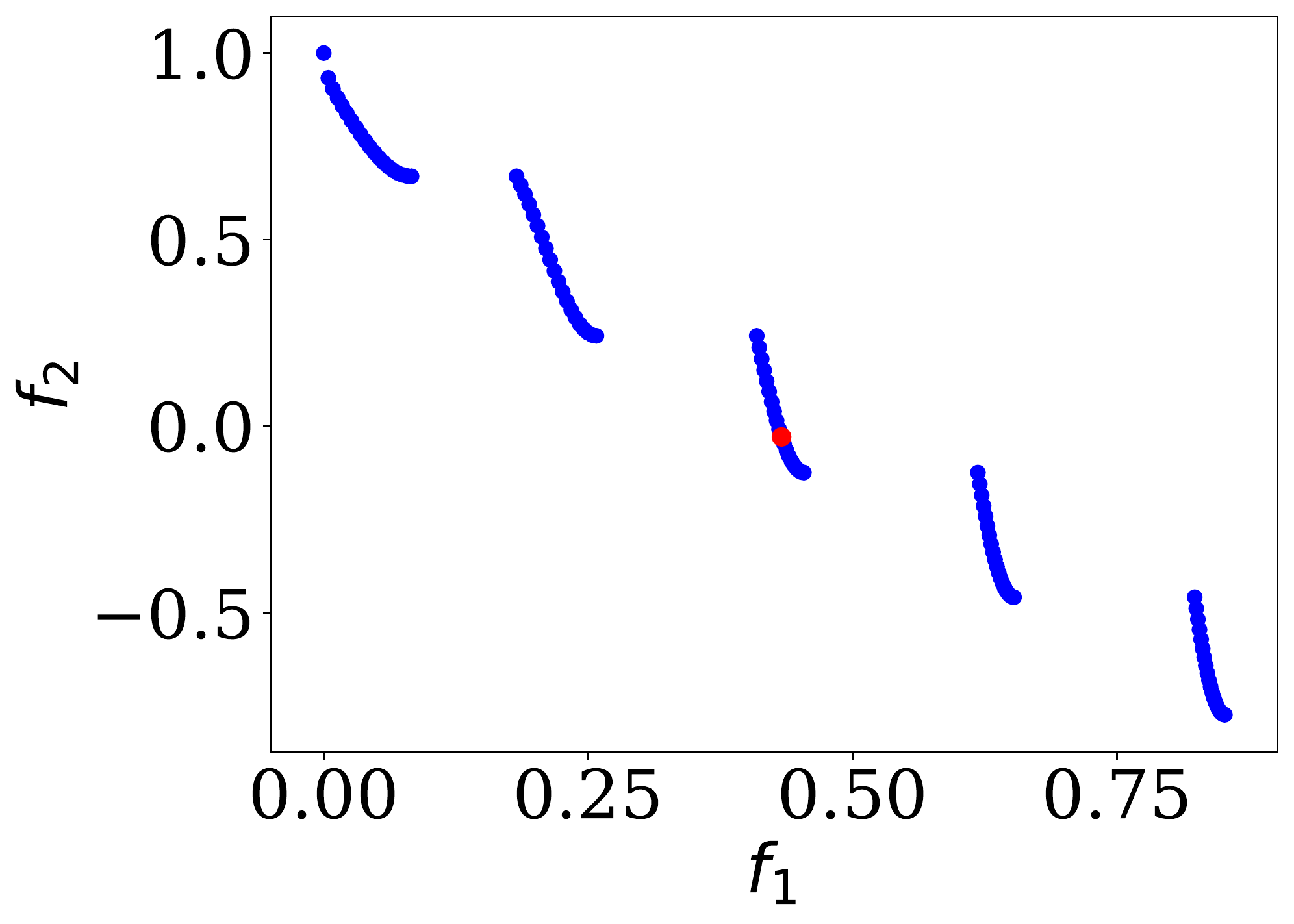}}
    \hfill
  \subfloat[Scatter 3D\label{fig:viz_scatter3d}]{%
        \includegraphics[width=0.3\linewidth]{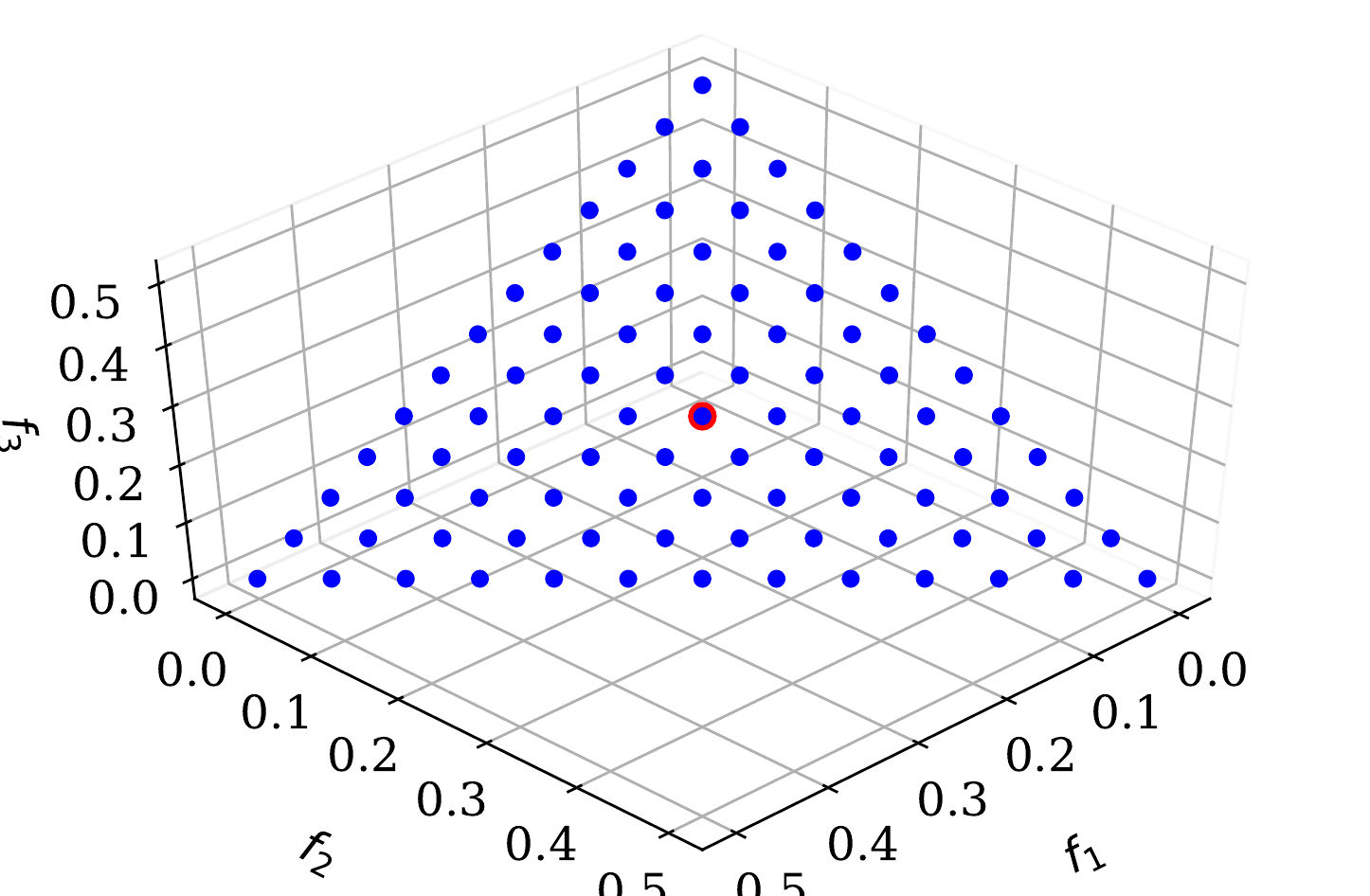}}
    \hfill
  \subfloat[Scatter ND\label{fig:viz_scatter_pairwise}~\cite{scatter_pairwise}]{%
        \includegraphics[width=0.3\linewidth]{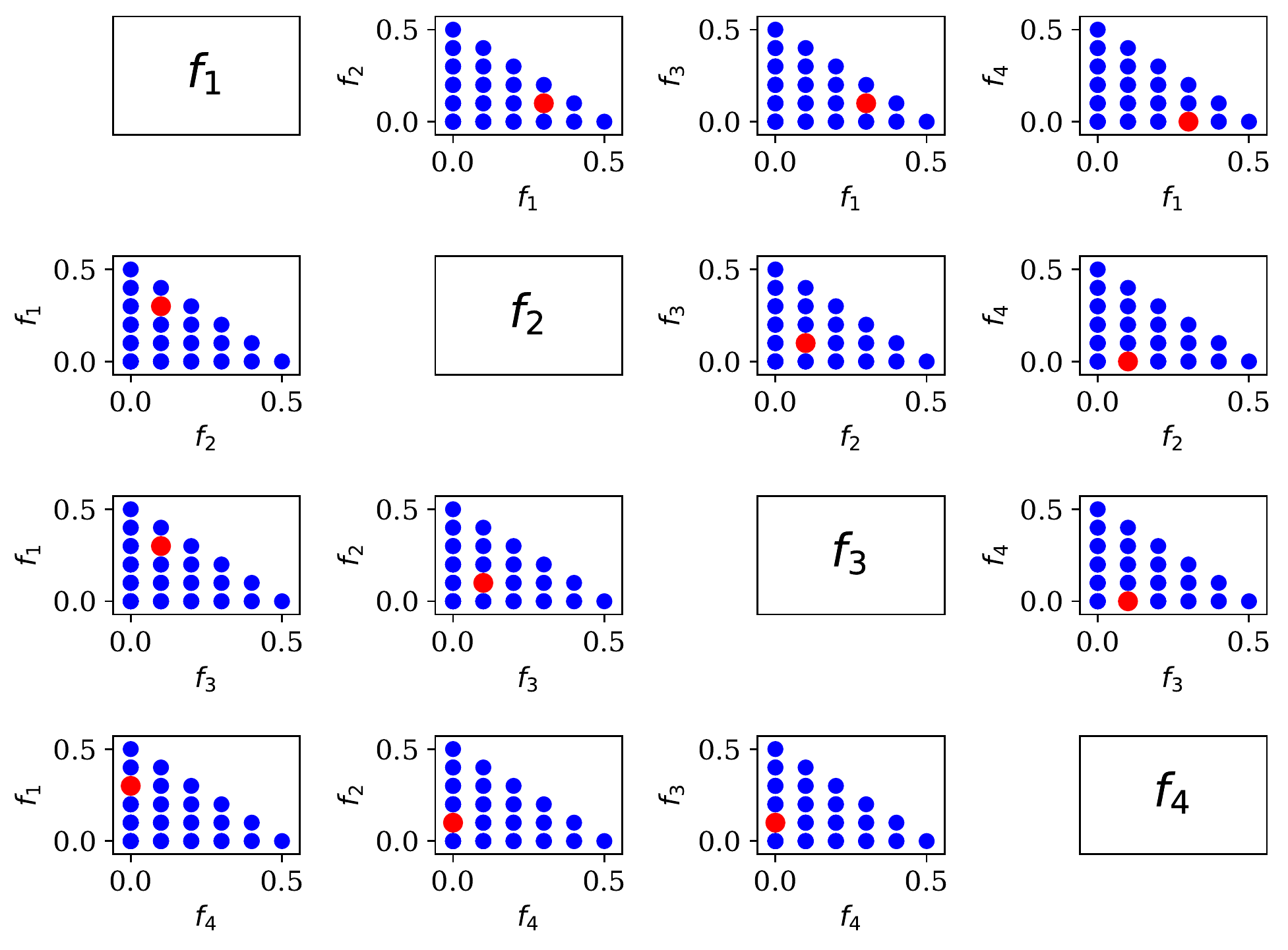}}
    \\
      \subfloat[PCP\label{fig:viz_pcp}~\cite{pcp}]{%
        \includegraphics[width=0.3\linewidth]{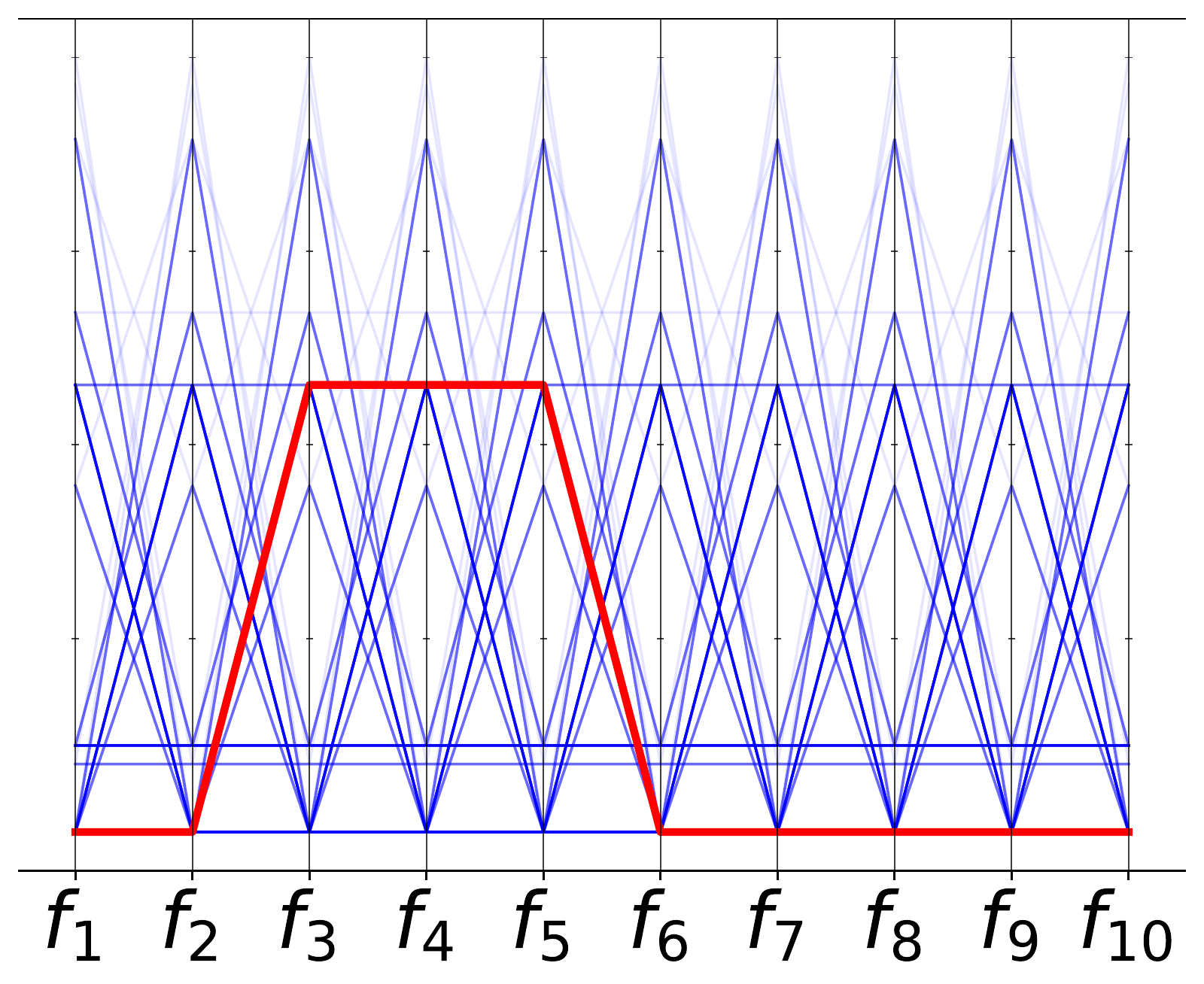}}
    \hfill
      \subfloat[Radviz\label{fig:viz_radviz}~\cite{radviz}]{%
        \includegraphics[width=0.3\linewidth]{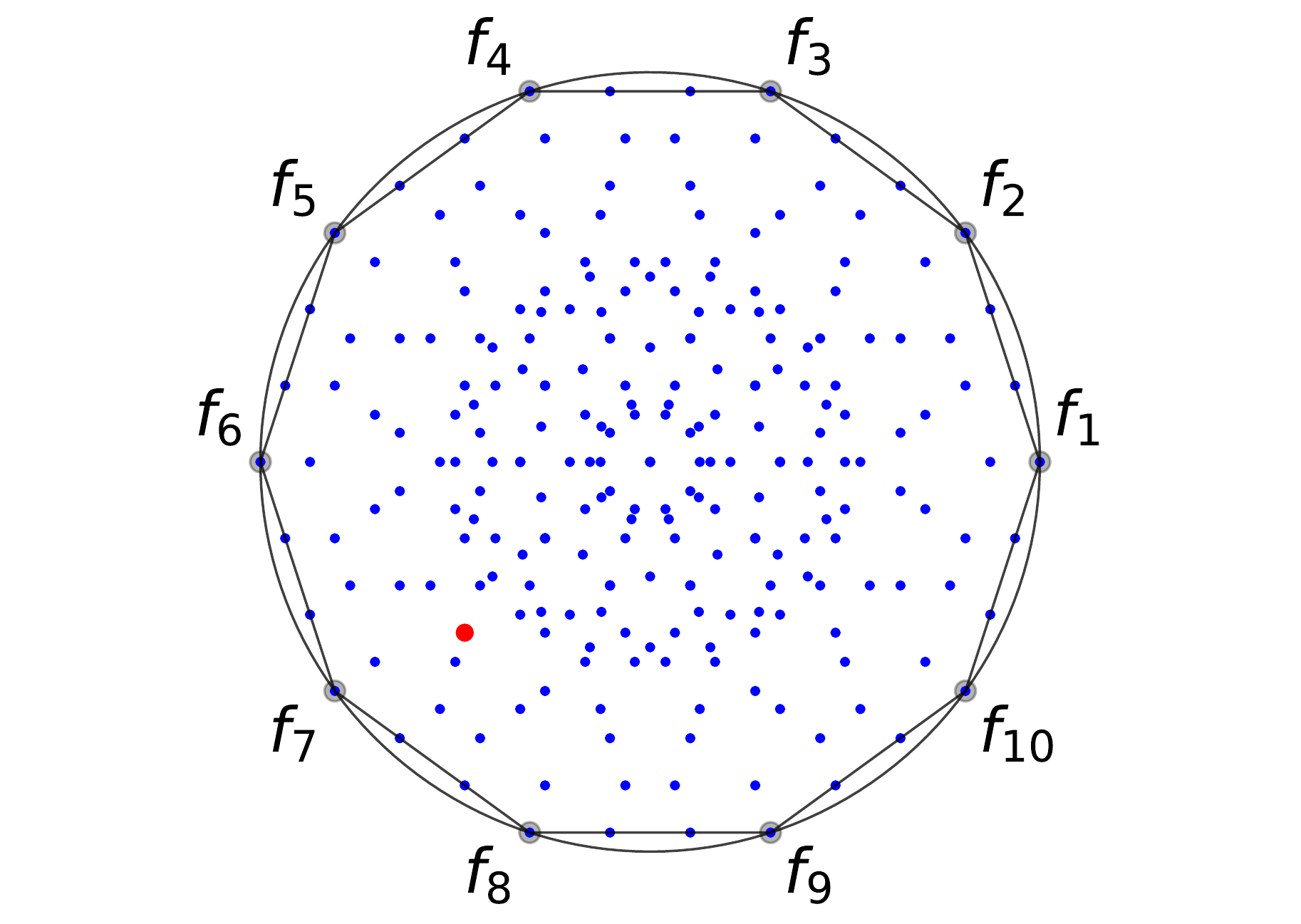}}
    \hfill
  \subfloat[Star Coordinate Graph~\cite{star}\label{fig:viz_star}]{%
        \includegraphics[width=0.3\linewidth]{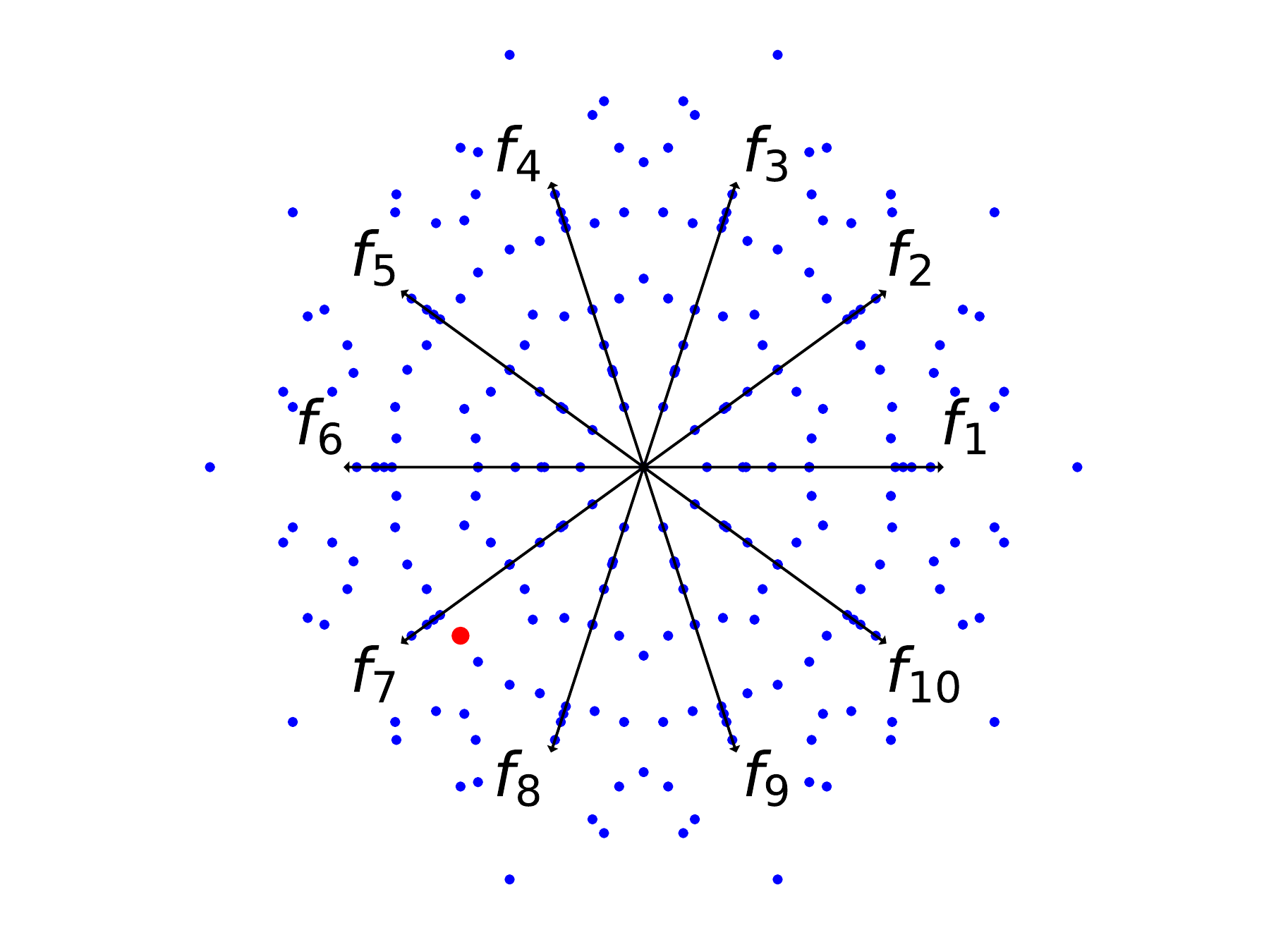}}
            \\
  \subfloat[Heatmap\label{fig:viz_heatmap}~\cite{heatmap}]{%
        \includegraphics[width=0.3\linewidth]{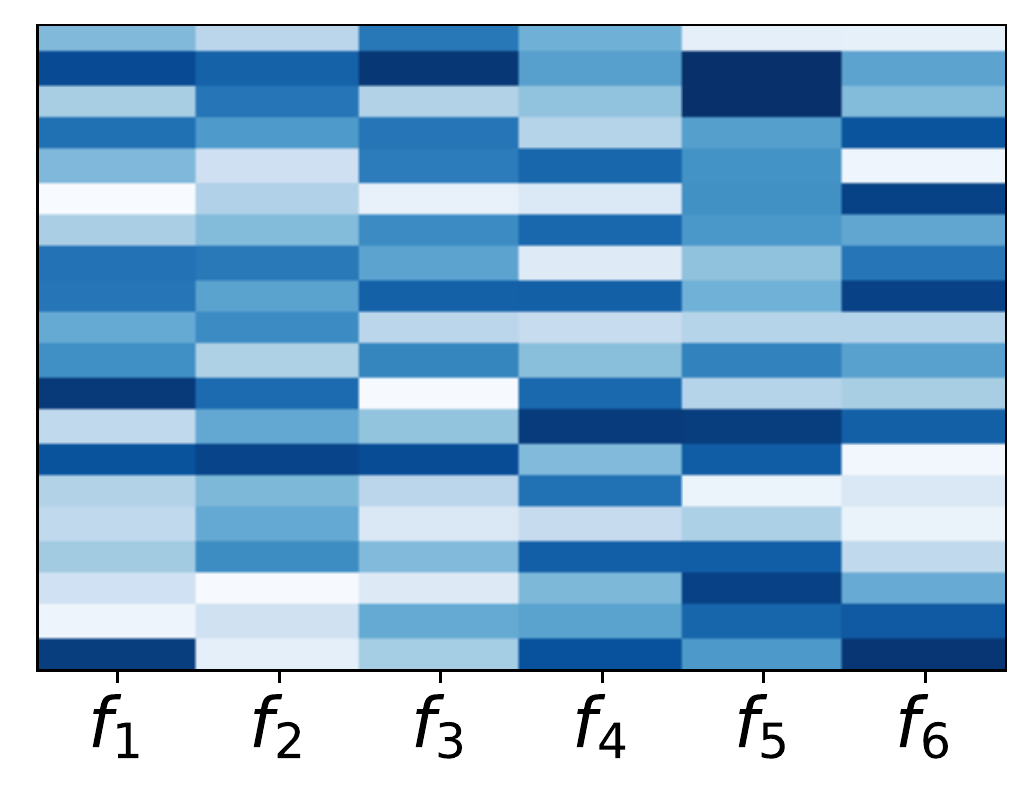}}
    \hfill
  \subfloat[Petal Diagram\label{fig:viz_petal}~\cite{petal}]{%
        \includegraphics[width=0.3\linewidth]{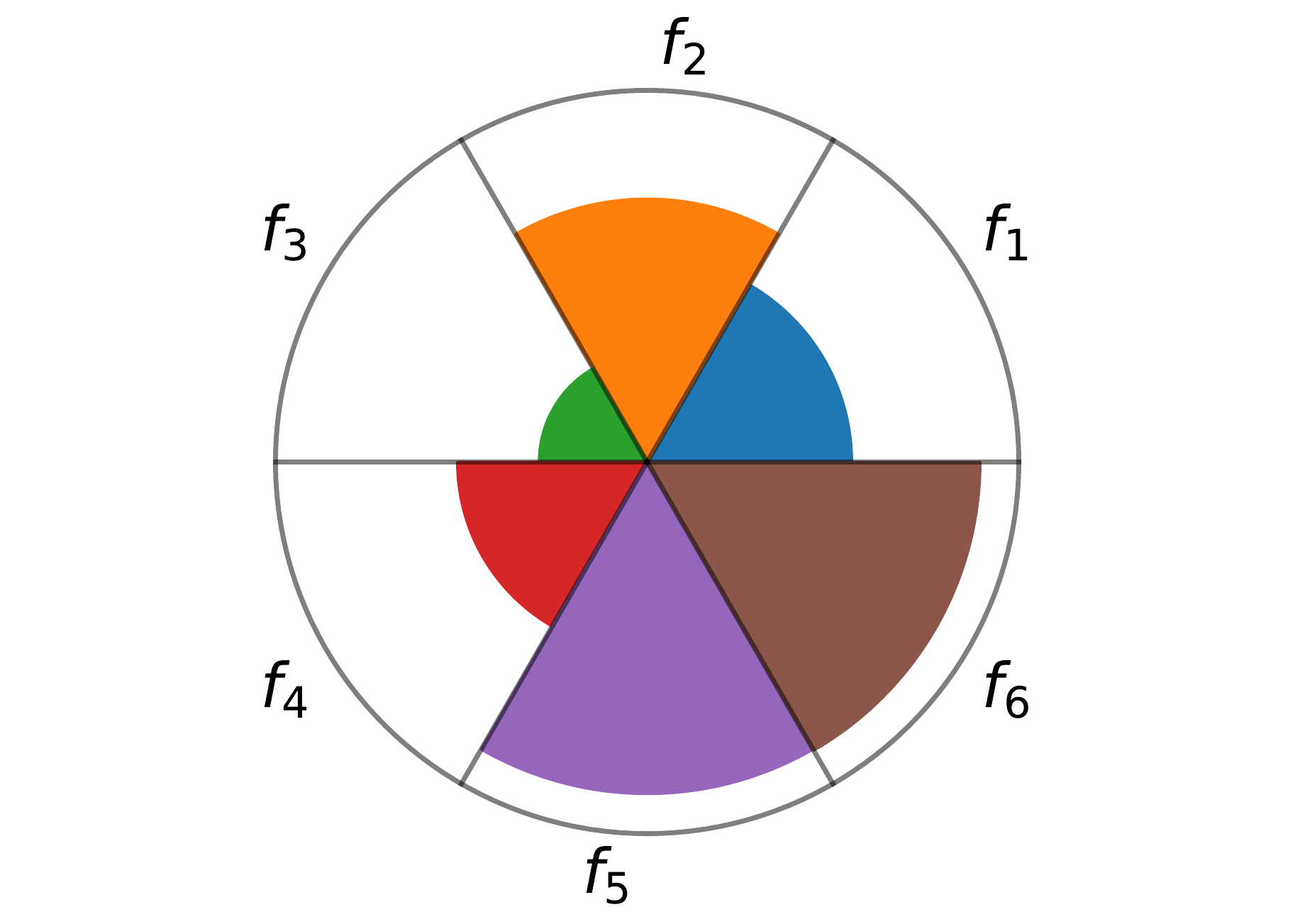}}
    \hfill
  \subfloat[Spider-Web/Radar\label{fig:viz_radar}~\cite{radar}]{%
        \includegraphics[width=0.3\linewidth]{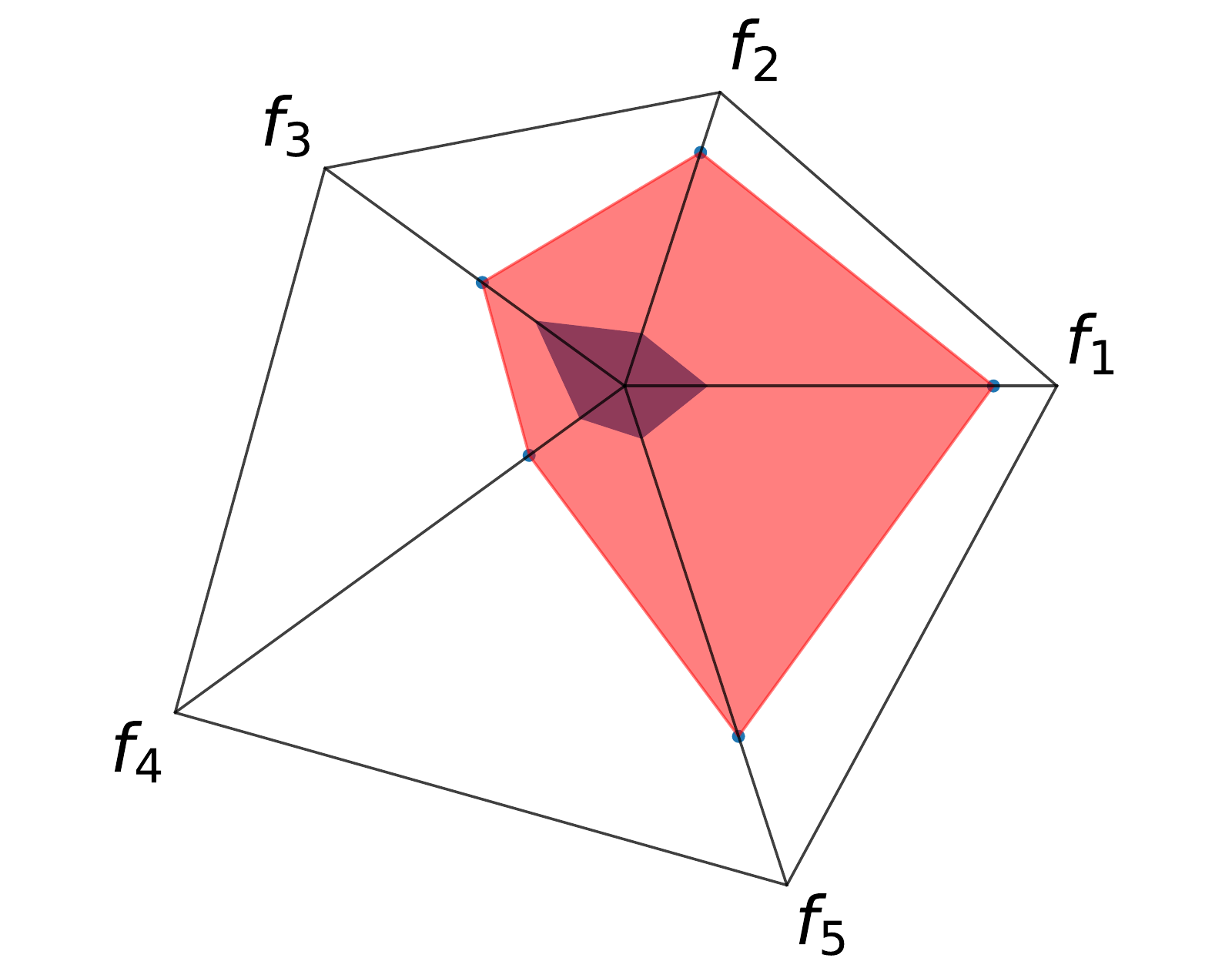}}
  \caption{Different visualization methods coded in {\tt pymoo}.}
  \label{fig:visualization} 
\end{figure*}

For $2$ or $3$ objectives, scatter plots (see Figure~\ref{fig:viz_scatter} and \ref{fig:viz_scatter3d}) can give a good intuition about the solution set. Trade-offs can be observed by considering the distance between two points. It might be desired to normalize each objective to make sure a comparison between values is based on relative and not absolute values. Pairwise Scatter Plots (see Figure~\ref{fig:viz_scatter_pairwise}) visualize more than $3$ objectives by showing each pair of axes independently. The diagonal is used to label the corresponding objectives.

Also, high-dimensional data can be illustrated by Parallel Coordinate Plots (PCP) as shown in Figure~\ref{fig:viz_pcp}. All axes are plotted vertically and represent an objective. Each solution is illustrated by a line from the left to the right. The intersection of a line and an axis indicate the value of the solution regarding the corresponding objective. For the purpose of comparison solution(s) can be highlighted by varying color and opacity.

Moreover, a common practice is to project the higher dimensional objective values onto the 2D plane using a transformation function. Radviz (Figure~\ref{fig:viz_radviz}) visualizes all points in a circle and the objective axes are uniformly positioned around on the perimeter.
Considering a minimization problem and a non-dominated set of solutions, an extreme point very close to an axis represents the worst solution for that corresponding objective, but is comparably "good" in one or many other objectives. Similarly, Star Coordinate Plots (Figure~\ref{fig:viz_star}) illustrate the objective space, except that the transformation function allows solutions outside of the circle.

Heatmaps (Figure~\ref{fig:viz_heatmap}) are used to represent the goodness of solutions through colors. Each row represents a solution and each column a variable. We leave the choice to the end-user of what color map to use and whether light or dark colors illustrate better or worse solutions. Also, solutions can be sorted lexicographically by their corresponding objective values.

Instead of visualizing a set of solutions, one solution can be illustrated at a time. The Petal Diagram (Figure~\ref{fig:viz_petal}) is a pie diagram where the objective value is represented by each piece's diameter. Colors are used to further distinguish the pieces. Finally, the Spider-Web or Radar Diagram (Figure~\ref{fig:viz_radar}) shows the objectives values as a point on an axis. The ideal and nadir point~\cite{multi_objective_book} is represented by the inner and outer polygon. By definition the solution lies in between those two extremes. If the objective space ranges are scaled differently, normalization for the purpose of plotting can be enabled and the diagram becomes symmetric.

\subsection{Decision Making}

In practice, after obtaining a set of non-dominated solutions a single solution has to be chosen for implementation.

\begin{enumerate}[label=(\roman*)]

\item \textbf{Compromise Programming:}
One way of making a decision is to compute value of a scalarized and aggregated function and select one solution based on minimum or maximum value of the function. In \texttt{pymoo} a number of scalarization functions described in Section~\ref{sec:decomp} can be used to come to a decision regarding desired weights of objectives.

\vspace{2mm}
\item \textbf{Pseudo-Weights}:
However, a more intuitive way to chose a solution out of a Pareto-front is the pseudo-weight vector approach proposed in~\cite{multi_objective_book}. The pseudo weight $w_i$ for the $i$-th objective function is calculated by:

\begin{equation}
w_i = \frac{(f_i^{\max} - f_i {(x)}) \, /\,  (f_i^{\max} - f_i^{\min})}{\sum_{m=1}^M (f_m^{\max} - f_m (x)) \, /\,  (f_m^{\max} - f_m^{\min})}. 
\end{equation}

The normalized distance to the worst solution regarding each objective $i$ is calculated. It is interesting to note that for non-convex Pareto-fronts, the pseudo weight does not correspond to the result of an optimization using the weighted sum method.

\vspace{2mm}
\item \textbf{High Trade-Off Solutions:}
Furthermore, high trade-off solutions are usually of interest, but not straightforward to detect in higher-dimensional objective spaces. We have implemented the procedure proposed in~\cite{knee_point}.
It was described to be embedded in an algorithm to guide the search; we, however, use it for post-processing. The metric for each solution pair $x_i$ and $x_j$ in a non-dominated set is given by:

\begin{equation}
T(x_i, x_j) = \frac{\sum_{i=1}^M \max[0,f_m(x_j) - f_m(x_i) ]} {\sum_{i=1}^M \max[0,f_m(x_i) - f_m(x_j)]},
\end{equation}
where the numerator represents the aggregated sacrifice and the denominator the aggregated gain.
The trade-off measure $\mu(x_i, S)$ for each solution $x_i$ with respect to a set of solutions $S$ is obtained by:

\begin{equation}
\mu(x_i, S) = \min_{x_j \in S} T(x_i, x_j)
\end{equation}

It finds the minimum $T(x_i, x_j)$ from $x_i$ to all other solutions $x_j \in S$. Instead of calculating the metric with respect to all others, we provide the option to only consider the $k$ closest neighbors in the objective space to reduce the computational complexity.
\end{enumerate}

Multi-objective frameworks should include methods for multi-criteria decision making and support end-user further in choosing a solution out of a trade-off solution set. 

\section{Concluding Remarks}
\label{sec:conclusion}

This paper has introduced \texttt{pymoo}, a multi-objective optimization framework in Python. We have walked through our framework beginning with the installation up to the optimization of a constrained bi-objective optimization problem. Moreover, we have presented the overall architecture of the framework consisting of three core modules: Problems, Optimization, and Analytics. Each module has been described in depth and illustrative examples have been provided. We have shown that our framework covers various aspects of multi-objective optimization including the visualization of high-dimensional spaces and multi-criteria decision making to finally select a solution out of the obtained solution set. One distinguishing feature of our framework with other existing ones is that we have provided a few options for various key aspects of a multi-objective optimization task, providing standard evolutionary operators for optimization, standard performance metrics for evaluating a run, standard visualization techniques for showcasing obtained trade-off solutions, and a few approaches for decision-making. Most such implementations were originally suggested and developed by the second author and his collaborators for more than 25 years. Hence, we consider that the implementations of all such ideas are authentic and error-free. Thus, the results from the proposed framework should stay as benchmark results of implemented procedures. 

However, the framework can be extended to make it more extensive. In the future, we plan to implement a more optimization algorithms and test problems to provide more choices to end-users. Also, we aim to implement some methods from the classical literature on single-objective optimization which can also be used for multi-objective optimization through decomposition or embedded as a local search. So far, we have provided a few basic performance metrics. We plan to extend this by creating a module that runs a list of algorithms on test problems automatically and provides a statistics of different performance indicators.

Furthermore, we like to mention that any kind of contribution is more than welcome. We see our framework as a collaborative collection from and to the multi-objective optimization community. By adding a method or algorithm to \texttt{pymoo} the community can benefit from a growing comprehensive framework and it can help researchers to advertise their methods. 
In general, different kinds of contributions are possible and more information can be found online. Moreover, we would like to mention that even though we try to keep our framework as bug-free as possible, in case of exceptions during the execution or doubt of correctness, please contact us directly or use our issue tracker.




\bibliographystyle{cas-model2-names}

\bibliography{cas-dc}




\end{document}